\documentclass[conference]{IEEEtran}
\IEEEoverridecommandlockouts
\usepackage{cite}
\usepackage{amsmath,amssymb,amsfonts}
\usepackage{algorithmic}
\usepackage{algorithm}
\usepackage{graphicx}
\usepackage{url}
\usepackage{textcomp}
\usepackage{xcolor}
\usepackage{caption2}
\usepackage{subfigure}
\usepackage{float}
\usepackage{bm}
\def\BibTeX{{\rm B\kern-.05em{\sc i\kern-.025em b}\kern-.08em
    T\kern-.1667em\lower.7ex\hbox{E}\kern-.125emX}}

\begin{document}

\title{Context-Based Soft Actor Critic for Environments with Non-stationary Dynamics\\
}


\author{\IEEEauthorblockN{Yuan Pu,
		Shaochen Wang, Xin Yao, Bin Li }
	\IEEEauthorblockA{University of Science and Technology of China, Hefei, China\\
		Email: \{puyuan, samwang, xinyao\}@mail.ustc.edu.cn,
		binli@ustc.edu.cn}}

\maketitle
\begin{abstract}
    The performance of deep reinforcement learning
    methods prone to degenerate when applied to environments with non-stationary
    dynamics. In this paper, we utilize the latent context
    recurrent encoders motivated by recent Meta-RL materials, and
    propose the Latent Context-based Soft Actor Critic (LC-SAC)
    method to address aforementioned issues. By minimizing the
    contrastive prediction loss function, the learned context variables
    capture the information of the environment dynamics and the
    recent behavior of the agent. Then combined with the soft policy
    iteration paradigm, the LC-SAC method alternates between
    soft policy evaluation and soft policy improvement until it
    converges to the optimal policy. Experimental results show that
    the performance of LC-SAC is significantly better than the SAC
    algorithm on the MetaWorld ML1 tasks whose dynamics changes
    drasticly among different episodes, and is comparable to SAC on
    the continuous control benchmark task MuJoCo whose dynamics
    changes slowly or doesn’t change between different episodes.
    At the same time, we also conduct relevant experiments to
    determine the impact of different hyperparameter settings on the
    performance of the LC-SAC algorithm and give the reasonable
    suggestions of hyperparameter setting.
\end{abstract}

\section{Introduction}
Recent years, deep reinforcement learning (DRL) algorithms \cite{b1} have present many impressive results in many challenging sequential-decision and control domains, such as Atari 2600 video games \cite{b2}, board games \cite{b3}, robot manipulation tasks \cite{b4} \cite{b5} . 

To achieve high sample efficiency and robust performance, many works have been advanced. Especially, soft actor-critic algorithm \cite{b6} achieves state-of-the-art performance on many challenging continuous control benchmarks, which is an off policy actor-critic setting RL algorithm based on the maximum entropy reinforcement learning framework. 
The actor reuses past experience through off-policy optimization and acts maximize expected reward while also maximizing entropy to increase exploration.

However, when one task implicitly requires long-horizon sequential decisions or varies with initial conditions or specific dynamics, the conventional algorithms usually could not solve it because the agent lacks effective memory mechanism and cannot obtain the necessary long-horizon information. 
For example, in MetaWorld ML1 task \cite{todorov2012mujoco}, Push-v2, the simulated mechanical arm aims to push a red-brown ice hockey into a target location. The initial location and target location changes randomly every episode. 
In such environment, whether the agent can accurately estimate the dynamics of the current environment and have a long-term memory of the movement trajectory of the robotic arm is very important.
A natural improvement could be explicitly increase the agent's memory length. In previous research \cite{b7}, their proposed Deep Recurrent Q-Network (DRQN), successfully integrates information through time and replicates DQN’s performance on discrete Atari domains. However through empirical experiments, they found that using recurrent neural networks can not guarantee systematic performance improvement.

Recently, in Meta-RL domain, \cite{b8} \cite{b9} \cite{b10} have proposed to encode task's salient identification information to a latent embedding space to reuse experience among different similar tasks and achieve both meta training and adaptation efficiency.

Motivated by these insights, in this paper, we proposed a latent context based soft actor critic method (LC-SAC).  In this method, an additional latent context encoder module is indroduced. The encoder utilizes the recurrent neural network structure, and take triples (state, action and reward) as inputs, and the context variable as outputs. 
The latent context vector encoder is updated by minimizing the contrative prediction loss function, so that the learned context vector captures the information of the environment dynamics and the recent behavior of the agent, which is very helpful for robust and effective policy update in the environment with non-stationary dynamics.  Experimental results show that its performance is comparable to SAC on the continuous control benchmark task MuJoCo with less obvious dynamic changes, and its performance is significantly better than the SAC algorithm on the MetaWorld ML1 task with drastic dynamic changes.

\section{Related works}

\subsection{Soft Actor-Critic}
Before introducing SAC, we first present the deep reinforcement learning problem definition. Reinforcement learning problem is often formulated as a {\em Markov Decision Process} (MDP), $\mathcal{M}=\left(\mathcal{S}, \mathcal{A}, p, r, \gamma\right)$. When the RL agent interacting with the environment, at each step, the agent observes a state ${\mathbf{s}_{t} \in \mathcal{S}}$, where $\mathcal{S}$ is the state space, and chooses an action ${\mathbf{a}_{t}\in\mathcal{A}}$, according to policy $\pi(\mathbf{a}_{t}|\mathbf{s}_{t})$, where $\mathcal{A}$ is the state space, then the agent receives a reward $r\left(\mathbf{s}_{t}, \mathbf{a}_{t}\right)$ and the environment transforms to next state $\mathbf{s}_{t+1}\sim p(\mathbf{s}_{t+1}|\mathbf{s}_{t},\mathbf{a}_{t})$. Note that the environment's dynamics is contained in both the transition probabilities $p$ and reward functions $r$, which are usually unknown to the agent.

The objective of maximum entropy reinforcement learning framework \cite{b15} is to maximize the discounted expected total reward plus the expected entropy of the policy:
\begin{equation}
J(\pi)=\sum_{t=0}^{T} \mathbf{E}_{\left(\mathbf{s}_{t}, \mathbf{a}_{t}\right) \sim \rho_{\pi}}\left[r\left(\mathbf{s}_{t}, \mathbf{a}_{t}\right)+\alpha \mathcal{H}\left(\pi\left(\cdot | \mathbf{s}_{t}\right)\right)\right]
\end{equation}
, where $\gamma $ is the discounted factor, $\rho_{\pi}\left(\mathbf{s}_{t},\mathbf{a}_{t}\right)$ denotes the state-action marginal of the trajectory distribution induced by the policy $\pi(\mathbf{a}_{t}|\mathbf{s}_{t})$.

{\em Soft Actor-Critic} (SAC) \cite{b6} is an off-policy actor-critic method
in the maximum entropy reinforcement learning framework. It utilizes an actor-critic architecture with separate policy and value networks, an off-policy paradigm that enables reuse of previously collected data, and entropy maximization to enable effective exploration. In contrastive to other off-policy algorithms, SAC is quite stable and achieving state-of-the-art results on a range of continuous control benchmarks. To verify our hypothesis
that latent context could increase the the agent's memory length and implicitly benefit the representation learning, our modification is based on SAC.

\subsection{RL with Recurrent Neural Networks}
One way to tackle memory-required problems is to use recurrent neural networks. Many prior work \cite{b7} \cite{b16} \cite{b17} have combined LSTM networks with policy gradient methods to solve POMDPs. \cite{b7} investigated the effects of adding recurrence to a Deep Q-Network (DQN), successfully integrates information through time and receives comparable scores with DQN on most standard Atari games.
Meanwhile, \cite{b16} proposed to extend deterministic policy gradient and stochastic value gradient to recurrent neural networks variants to solve challenging memory problems such as the Morris water maze. 

Recently, \cite{b17} empirically investigated the training of RNN-based RL agents from distributed prioritized experience replay, and obtained remarkably good results in Atari games domain. But it also showed representational drift and recurrent state staleness problem is exacerbated in the continuous training setting, and ultimately results in diminished training stability and performance.
Meanwhile in \cite{b8}, experimentally they found that straightforward incorporation of recurrent policies with off-policy learning is difficult, especially in continuous settings. 

\subsection{Probabilistic Meta-RL}
In Meta-RL materials, many approaches \cite{b18} \cite{b19} \cite{b20} have been proposed to transfer among different similar tasks. MAESN \cite{b19} combines structured stochasticity with MAML \cite{b18} by learning exploration strategies from prior experience, resulting in fast adaptation to new tasks.

The most related work with us is PEARL \cite{b8} algorithm. They represent task contexts with probabilistic latent variables, which encode the commonness of different tasks in some meaning. This probabilistic interpretation enables temporally extended exploration behaviors that enhance adaptation efficiency while requiring far less experience. 

Our proposed LC-SAC algorithm is a variant based on PEARL. The main difference is that, PEARL aims to solve Meta-RL problems, so the latent context variables $\mathbf{c}$ encode salient identification information about the task, while in our LC-SAC, the latent context is trained to memorize the recent information about the dynamics of the current undergoing environment and the agent's past actions (or behaviors).

Our main concern in this paper is if adding latent context will impair conventional RL algorithms' performance in simple continuous tasks in different experimental parameters settings and if latent context can be used to improve original SAC algorithms' performance in tasks with non-stationary dynamics




\section{Methods}

It is worth noting that the latent context encoder module proposed can be flexibly combined with various basic reinforcement learning algorithms. In this paper, the SAC algorithm is used as the basic algorithm, which is the SOTA  method in the current continuous control domain. And the following algorithm description and experimental research are carried out based on SAC. We call the new algorithm obtained as Latent Context Based Soft Actor Critic, which is denoted as LC-SAC.
In this section, we first introduces the basic principles of the latent context encoder and the contrastiveive predictive loss function that used to update the encoder network. Then we introduces how to effectively combine the obtained context variables with the soft policy iteration paradigm.


\subsection{Latent Context Encoder}
In the actual test environment, the dynamics of the environment (including the state transition function $r(s_t,a_t)$ and the reward function $P(s_{t+1}|s_t,a_t)$)
 is usually unknown, but the trajectory obtained by the agent interacting with the environment contains relevant information, and the characterization of environmental dynamics can be learned from these trajectory samples. The triplet composed of the state, action, and reward collected by the agent at the time step $t$ is called the transition experience, which is denoted as $e_t=(s_t,a_t,r_t)$, and we use
$e_{1:H}$ 
represents the experience collected in the last H time steps, where H represents the time step Horizon. The latent context encoder network in the LC-SAC algorithm is denoted as $q_{\omega}(c|e)$, which is a mapping from $\mathbf{e} \in \mathcal{E}$ to $\mathbf{c} \in \mathcal{C}$, where $\mathcal{E}$ represents the experience space, $\mathcal{C}$ represents the latent context vector space, and $\omega$ represents the parameters of the latent context encoder neural network.

The latent context encoder adopts the recurrent neural network (LSTM) structure. At each time step t, the input of the encoder are the triplet consisting of state, action and reward $e_t$ and the hidden state of the previous time , and the output is the latent context vector $e_t$. The algorithm sample the continuous fragments of the trajectory from the replay buffer $\mathcal{D}_{c}$, and through the stochastic gradient descent method, minimize a contrastiveive prediction loss function \cite{oord2018representation} to update the network parameters of the context encoder, which can be proved equivalent to maximizing the mutual information between the context variable $c_t$ at time t and the triplet composed of state, action, and reward at the next time t+1.
This operation will make the learned context vector as a kind of implicit representation of the continuous sequence fragments of the trajectory, and contain information on environmental dynamics and the recent behavior of the agent $\pi(a_t|s_t)$.
The contrastiveive prediction loss function comes from the InfoNCE (Information Noise Contrastive Estination) loss function in the work of contrastive learning \cite{oord2018representation}. In the following, we first gives the principle of the InfoNCE loss function and how to apply it to the learning of latent context variables, and then introduces the structural design and update process of the latent context encoder.

\subsubsection{InfoNCE Loss}
First, we give the definition of mutual information. For two random variables $c$ and $e$, the mutual information (MI) between them is defined as:
\begin{equation}
I(e; c)=\sum_{e, c} p(e, c) \log \frac{p(e \mid c)}{p(e)}
\end{equation}
In order to model the probability dependence relation between $c_t$ (context variable at the current time step) and $e_{t+1}$ (that is, the transition triplet composed of the state, action and reward at the next time step)
Directly using the conditional generation model $p(e|c)$ may not be the best, because when modeling $p(e|c)$ directly, the algorithm tends to pay more attention to reconstruct details of the data, however, which requires a lot of calculations ability, and usually make the model ignore the contextual information among a long time span but this is required for effective optimization of the policy.

If we do not directly use the generative model $p_k (e_(t+k) |c_t)$ to predict the transition after k time steps in the future $e_(t+k) $
, \cite{oord2018representation} 
pointed out that we can utilize the joint probability density of variables $e_{t+k}$ and $c_{t}$, 
and the relation between joint probability density and the generative model $p_k (e_{t+k} | c_t)$
are the following expressions:
\begin{equation}
f_{k}\left(e_{t+k}, c_{t}\right) \propto \frac{p\left(e_{t+k} \mid c_{t}\right)}{p\left (e_{t+k}\right)}
\end{equation}

In this paper, we use a simple log-bilinear model to represent the joint probability density function, namely
$ f_{k}\left(e_{t+k}, c_{t}\right)=\exp \left(e_{t+k} W_{k} c_{t}^{T}\right)$ .
Although it is not possible to directly learn the probability distribution function of p(e) or p(e|c), but we can use samples from these distributions, so that we can use Noise contrastiveive Estination (NCE) technology\cite{2010Noise} It is mainly based on comparing target values with randomly sampled negative values to learn the joint probability density function of several variables.

In our implementation, the LC-SAC algorithm uses the InfoNCE loss function proposed by \cite{oord2018representation}.
First, given a set $E = {e_1, …, e_N}$, it contains a total of $N$ random samples, of which only the sample sampled from the distribution $p(e_{t+k} |c_t)$ is positive sample, the other N-1 samples are negative samples from the proposal distribution $p(e_{t+k})$,
the specific calculation formula of the InfoNCE loss function \cite{oord2018representation} is:
\begin{equation}
\mathcal{L}_{CP}^N=-\underset{E}{\mathbb{E}}\left[\log \frac{f_{k}\left(e_{t+1}, c_{t }\right)}{\sum_{e_{j} \in E} f_{k}\left(e_{j}, c_{t}\right)}\right]
\end{equation}
In the subsequent experiments in this paper, the LC-SAC algorithm updates the context encoder by optimizing the above InfoNCE loss function, taking k=1, that is, learning the latent context variables by predicting the transition experience at the next time step. In the realization of the formula (4), first sample a mini-batch containing N samples. Each sample is a transition sequence of length 20, which is a continuous segment of the trajectory. For each sample, the context variable $c_t$  generated at time t after passing through the context encoder, the transition $e_{t+1}$ of other N-1 samples at time $t+1$ are negative values, and the transition of this sample at time $t+1$, $e_{t+1}$ is a positive value.

In fact, optimizing this loss function will cause $f_k(e_{t+k},c_t )$ to estimate the density ratio in formula (3).
After theoretical derivation \cite{oord2018representation}, it can be obtained that the relationship between the above-mentioned InfoNCE loss function and the mutual information between $e_{t+k}$, $c_{t}$ is as follows:
$I\left(e_{t+k}, c_{t}\right) \geq \log (N)-\mathcal{L}_{\mathrm{CP}}^N$
. The above formula shows that minimizing the InfoNCE loss $\mathcal{L}_{\mathrm{CP}}^N$ can maximize the lower bound of the mutual information between the context variable at the current time step and the transition variable at the next time step, thereby implicitly maximize the mutual information between them.

\subsubsection{Structure and Training Pipline of Latent Context Encoder}
Context variables have two forms of expression, one is a deterministic value, and the other is a probabilistic value. If the context variable is modeled as a probability distribution, denoted as $q_{\omega}\left(\mathbf{c} | \mathbf{e}\right)$, where $\omega$ represents the network parameters of the context encoder. Its network structure diagram is shown in Figure 1. If the context variable is modeled as a deterministic value, it is denoted as $\mathbf{c}=q_{\omega}\left( \mathbf{e}\right)$, and its network structure can be obtained with a few modifications to Figure 1. The only modification is to remove the linear layer represented by Linear logstd in the last part of the network structure, and retain a network header represented by Linear mu, then the output is the deterministic context. Since the determined value can be regarded as a special case of the probability value, for the sake of brevity, in the following formula, the context encoder is all marked as $q_\omega(c|e)$.

In order to learn the context encoder network $q_\omega(c|e) $. The overall optimization objective function is as follows:
\begin{equation}
J_\mathbf{q} \left(\omega\right) = \mathop{\mathbb{E}}_{\mathbf{c} \sim q_{\omega}\left(\mathbf{c} | \mathbf{ e}\right), \mathcal{D}_{c}}\left [\mathcal{L}_{CP}+ \beta_1 \mathcal{L}_{critic} +\beta_2
D_{\mathrm{KL}}\left(q_{\omega}\left(\mathbf{c} | \mathbf{e}\right) \| p(\mathbf{c})\right)\right]
\end{equation}

Here, $\mathcal{D}_{c}$ refers to  
the replay buffer:
$\left( \mathbf{s}_{t}, \mathbf{a}_{t},\mathbf{r}_{t}, \mathbf{s}_{t+1}\right) \sim \mathcal{D}_{c}$
. The context vector c is sampled from
the distribution $q_{\omega}\left(\mathbf{c} | \mathbf{e}\right)$  corresponding to the current context variable encoder network parameters.

In the formula (5), the first term is the contrastive prediction loss function expressed by the equation (4), which causes the learned context vector to capture the information of the environment dynamics and the recent behavior of the agent.
The second term is the conventional SAC value loss function, which is equivalent to a regularization term, forcing the learned context vector to not change too much from the previous iteration, which is conducive to the stability of context variable learning sometimes. In the experimental settings of this paper, specifically, 
$\mathcal{L}_{critic} = J_{Q}(\phi_1)+J_{Q}(\phi_2)$
The third item is KL divergence, so as to make, the probability distribution of the context variable is expected to be as close as possible to the Gaussian distribution with the mean value of 0 and the variance of the identity matrix. This draws on the idea of the variational autoencoder and sets such an information bottleneck., Which limits the exploration space of latent variables and makes the learned latent variables compact.
$\beta_1$,$\beta_2$ are hyperparameter factors that control the relative importance of these three items.
Note that items 2 and 3 are optional. In the experiments later in this paper, in most tasks, the context variable is a deterministic value, and the objective function only contains the first item.

When the algorithm is actually running, the agent will store the corresponding transition tuple $\left( \mathbf{s}_{t}, \mathbf{a}_{t},\mathbf{r}_ {t}, \mathbf{s}_{t+1}\right)$ in $\mathcal{D}_{c}$ once it interacts with the environment. When one episode is over (done=True), put the corresponding position index of the replay buffer into a list, and this list records the position of the end of each episode, which is convenient for sampling the continuous segment of the trajectory.
The sample used to learn the context encoder is to uniformly sample a  minibatch (each entry contains the continuous segment of the trajectory ) from $\mathcal{D}_{c}$.

\begin{figure*}[!htbp]
\centering
\includegraphics[width=14cm,height=3.5cm]{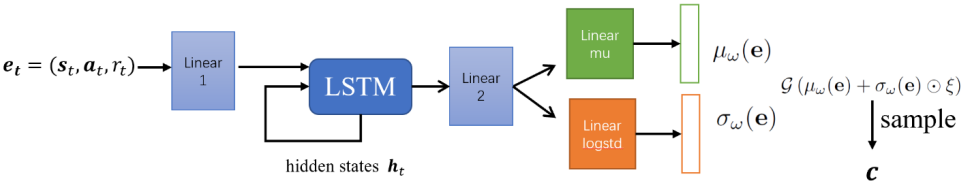}
\caption{Schematic diagram of latent context variable encoder network structure}
\end{figure*}

\begin{algorithm}[!htbp]
    \caption{LC-SAC algorithm}
    \label{algo:algorithm1}
        Initialize neural network parameters: latent context encoder network $\omega$, policy network $\theta$, action value network $\phi_{1,2}$, target action value network $\bar{\phi}_{1, 2}$;
        Initialize 2 replay buffers: $\mathcal{D}_{c}$ is used to train the latent context encoder, $\mathcal{D}_{rl}$ is used to train the policy network and the action value network;
        Given the maximum number of training steps $M$, context vector dimension $d$, used to update the sequence segment length $l$ of the context vector encoder, replay buffer size $rlbs$
        \begin{algorithmic}[1] 
        \FOR{$t \leftarrow 1$ to $M$ }
           	\STATE According to the trajectory currently experienced, $\mathbf{c} \sim q_{\omega}\left(\mathbf{c} | \mathbf{e}\right)$ is sampled by the context encoder,
            	Execution policy $\pi_{\theta}(\mathbf{a} | \mathbf{s}, \mathbf{c})$ interact with the environment to gain experience $\mathbf{e_t}$\;
           	\STATE Copy the sample $ \left( \mathbf{s}_{t}, \mathbf{a}_{t}, \mathbf{r}_{t}, \mathbf{s}_{t+1}\right) $ 	\STATE Add to replay buffer $\mathcal{D}_{c}$\;
           	\STATE Add sample $ \left( \mathbf{c_t}, \mathbf{s}_{t}, \mathbf{a}_{t},\mathbf{r}_{t}, \mathbf{s}_{t +1},\mathbf{c}^{\prime}\right)$ add to the replay buffer $\mathcal{D}_{rl}$\;
            \IF{t \% $N_{rl}$ = 0}
                \FOR{each rl training step}
                	\STATE Randomly sample a mini-batch uniformly from $\mathcal{D}_{rl}$ $\mathbf{B}_{rl}$\;
                	\STATE Respectively update the value network and policy network parameters according to the loss function (6), (8):
               \STATE  $\phi_{i} \leftarrow \phi_{i}-\alpha_{Q} \hat{\nabla}_{\phi_{i}} J_{Q}\left(\phi_{i}\right)$, for $i \in\{1,2\}$;
                $\theta \leftarrow \theta-\alpha_{\pi} \hat{\nabla}_{\theta} J_{\pi}(\theta)$\;
                	\STATE Update target value network parameters:
                $\bar{\phi}_{i} \leftarrow \tau \phi_{i}+(1-\tau) \bar{\phi}_{i}$, for $i \in {1,2} $\;
                 \ENDFOR
            \ENDIF
            \IF{t \% $N_c$ = 0}
            \FOR{each context training step}
                \STATE Randomly sample a {\bf continuous segment of the trajectory} mini-batch $\mathbf{B}_{c}$\;
                \STATE The trajectory segments obtained by sampling are sampled by the context encoder to obtain $\mathbf{c} \sim q_{\omega}\left(\mathbf{c} | \mathbf{e}\right)$ \;
                \STATE Update the network parameters of the context vector encoder according to the loss function (5):
                $\omega \leftarrow \omega-\alpha_{q} \nabla_{\omega} J_{q}(\omega)$\;
            \ENDFOR
            \ENDIF
   	\ENDFOR
    \STATE \textbf{return} Learned network parameters: $q_{\omega}, Q_{\phi_{1,2}}, \pi_{\theta}$ 
    \end{algorithmic}
\end{algorithm}

\subsection{Context-Based Soft Policy Iteration}

\begin{figure}[!htbp]
\centering
\includegraphics[width=8cm,height=9cm]{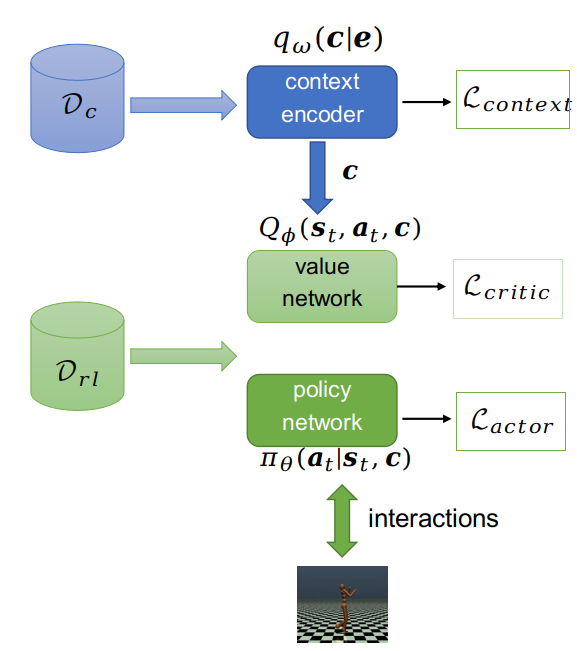}
\caption{Schematic diagram of LC-SAC algorithm training process}
\end{figure}

In this section, we introduces how to effectively combine the latent context variable encoder proposed in the previous section with the soft policy iteration paradigm to obtain our LC-SAC algorithm.

First we introduce the soft policy iteration.
In this paradigm, policy is represented  as $\pi_\theta$, where $\theta$ refers to the parameters of the neural network. The inputs of policy network are the state s and outputs is the executed action
(Or the probability distribution of the action). And we defines the soft action value function $Q^\pi_\phi(s,a)$, where $\phi$ refers to the parameters of the neural network, which is to evaluate the future cumulative discount reward plus the expected value of the policy entropy obtained by the agent after it perform actions according to the policy $\pi$ at any time after the state s.

The soft policy iteration generally adopts the stochastic gradient descent method to alternately optimize the value network and the policy network. 
The soft policy iteration paradigm is a reinforcement learning optimization framework utilizing the off-policy form . When updating the network, we can use the trajectory sequence generated by interacting with the environment using different policy since the start time. 
In the specific implementation, all collected samples are stored in a replay buffer, which is a first-in first-out queue. When the buffer is full, the old samples will be discarded and the new samples will be placed at the end of the queue. Because of the strong correlation between the continuous samples generated,
During experience replay, a small minibatch is uniformly sampled from the replay buffer to update the network. Such an experience replay mechanism can partially remove the correlation between samples, making the samples used for training more in line with machine learning's assumption of independent and identical distribution of samples. 

The LC-SAC algorithm uses 2 replay buffers, one of which stores $\left( \mathbf{s}_{t}, \mathbf{a}_{t},\mathbf{r}_{t}, \ mathbf{s}_{t+1}\right)$, denoted as $\mathcal{D}_{c}$, used to update the latent context encoder;
The other one stores  $\left( \mathbf{c}_{t},\mathbf{s}_{t}, \mathbf{a}_{t},\mathbf{r}_{t}, \ mathbf{c}_{t}^\prime,\mathbf{s}_{t+1}\right)$, denoted as $\mathcal{D}_{rl}$, used to update the value network and policy network .
Compared with the original SAC algorithm, the input of the value function network and the policy network of the LC-SAC algorithm has changed from the original state $s$ to the splicing of state and context variables, that is, concat($s,c$). The context variable $c$ is generated by the latent context variable encoder module proposed in the previous section.
Similar to the original SAC algorithm, LC-SAC method also uses two soft Q-value networks
$Q_{\phi_i}\left(\mathbf{s}_{t}, \mathbf{a}_{t},\mathbf{c} \right)$, for $i \in {1,2}$, and a policy network $\pi_{\theta}\left(\mathbf{a}_{t} | \mathbf{s}_{t},\mathbf{c}\right)$.
All networks add a latent context variable $\mathbf{c}$ to the original input $\mathbf{s}_t$,
This variable encodes recent experiences, including dynamics information of environments and agent’s past behavior information
, which implicitly make agent have a longer time memory.

For a given policy $\pi$, its state-action value can be updated for any state s, which is called soft policy evaluation, and is achieved by optimizing the following loss function:
\begin{equation}
J_{Q}(\phi_i)=\mathop{\mathbb{E}}_{
\begin{aligned}
\mathcal{D}_{rl}\\
\end{aligned}
}
\left[\frac{1}{2}\left(Q_{\phi_i}\left(\mathbf{c},\mathbf{s}_{t}, \mathbf{a}_{t}\right) -\hat{Q}_{\phi_i}\left(\mathbf{r}_{t},\mathbf{c^{\prime}},\mathbf{s}_{t+1}, \mathbf{ a}_{t+1}\right)\right)^{2}\right]
\end{equation}
where $\hat{Q}$ represents the target value to update the current Q network, specifically referring to the sum of the estimated value of the future cumulative discount reward obtained 
plus the entropy of the policy, coupled with the value of the reward $r_t$ obtained at the current time step, the specific calculation formula is as follows:
\begin{equation}
\begin{aligned}
\hat{Q}\left(\mathbf{r}_{t},\mathbf{c^{\prime}},\mathbf{s}_{t+1}, \mathbf{a}_{t+ 1}\right)
&=\mathbf{r}_{t}+\gamma E_{\left(\mathbf{s}_{t+1}, \mathbf{a}_{t+1}\right) \sim \rho_{ \pi}}\\
&\left[\bar{Q}\left(\mathbf{s}_{t+1}, \mathbf{a}_{t+1}, \mathbf{c}^{\prime}\right)
-\alpha \log \pi\left(\mathbf{a}_{t+1} \mid \mathbf{s}_{t+1}, \mathbf{c}^{\prime}\right)\right ]
\end{aligned}
\end{equation}
where $\bar{Q}$ is the target network of the soft action value. It has the same structure as the current Q neural network but has different parameters. Its parameters are the exponential moving average of the current Q network parameters.
This setting has been proven to make training more stable. $\alpha$ is a hyperparameter, which determines the relative importance between the entropy of the policy and the original reward, that is, controls the degree of exploration. $\mathcal{D}_{rl}$ contains {\bf context variable, state, action, reward, next state, next context variable} replay buffer: $\left( \mathbf{c}_{t},\mathbf{s} _{t}, \mathbf{a}_{t},\mathbf{r}_{t}, \mathbf{c}_{t}^\prime,\mathbf{s}_{t+1} \right) \sim \mathcal{D}_{rl}$.

Note that our experimental results show that  that if the $\mathbf{c}$ and $\mathbf{c}^{\prime}$ here are calculated by the current context encoder, it will make the algorithm unstable and difficult to converge.
 
The soft policy improvement is achieved by maximizing the current soft value function and the entropy of the policy. Specifically, 
the loss function of the policy network update is as follows:
\begin{equation}
J_{\pi}(\theta)=\mathop{\mathbb{E}}_{
\left(\mathbf{c}_{t},\mathbf{s}_{t}, \mathbf{a}_{t},\mathbf{r}_{t}\right)
\sim \mathcal{D}_{rl}}
\left[\alpha \log \pi_{\theta}\left(\mathbf{a}_t | \mathbf{s}_{t}, \mathbf{c} \right)-Q_{\phi}\left( \mathbf{s}_{t}, \mathbf{a}_t, \mathbf{c}\right)\right]
\end{equation}
Recent work has theoretically proved that the soft policy iteration can ensure the monotonic improvement of the policy and finally converge to the optimal policy.

The LC-SAC algorithm is divided into the following three stages, and the algorithm runs iteratively between these three stages. The details of the LC-SAC algorithm are summarized in Algorithm 1, and the overview training process is shown in Figure 2.

{\it Phase 1}: Collect data samples generated by interaction with the environment. At every time step, first obtain the context variable $c$ according to the context encoder, then execute actions following the current policy $\pi$, interact with the environment, get the experience transition, and store the transition into the replay buffer $\mathcal{D}_{c }$ and $\mathcal{D}_{rl}$;

{\it Phase 2}: Optimize the value network and policy network. Sample from replay buffer $\mathcal{D}_{rl}$ 
$\left( \mathbf{s}_{t}, \mathbf{a}_{t},\mathbf{r}_{t}, \mathbf{c},\mathbf{s}_{t+1},\mathbf{c}^{\prime}\right)$, update the value network and policy network parameters according to the loss function (6), (8) ;

{\it Phase 3}: Optimize the context encoder. Sample from replay buffer $\mathcal{D}_{c}$ 
$\left( \mathbf{s}_{t}, \mathbf{a}_{t},\mathbf{r}_{t}, \mathbf{s}_{t+1}\right)$, update the context encoder network parameters according to the contrastive prediction loss function (5).

The second stage includes two steps. The first step is soft policy evaluation, which is to continuously use soft Bellman operators to act on the Q network to evaluate the soft Q value of the current policy. The second step is  soft policy improvement. That is, the policy is optimized toward the exponential direction of the current soft action value. Recent work \cite{haarnoja2018soft} proves that alternately performing these two steps can make the final policy converge to the optimal policy in the sense of maximum entropy. Note that the input of all value networks and policy networks has changed from the original state $s$ to the concatenation vector $concat(s,c)$ of the state and context variables at the current time step, and the sapmles for updating them are from $\mathcal{D} _{rl}$ sampled in $\left( \mathbf{c}_{t},\mathbf{s}_{t}, \mathbf{a}_{t},\mathbf{r}_{t }, \mathbf{c}_{t}^\prime,\mathbf{s}_{t+1}\right)$.

The {\it Phase 2} and {\it Phase 3} of the LC-SAC algorithm are run alternately at certain time intervals.
The advantages of alternately training policy network and the context encoder are as follows: If a context encoder is pre-trained based on the samples generated by the random policy, the trajectory space explored by the random policy is limited, resulting in the sample set used to update the context encoder is biased.
If the policy network and the context encoder are trained alternately, since the policy is continuously improved during the learning process, after the policy network has learned a certain amount of knowledge, agent can collect good quality trajectories (usually measured by the cumulative score of one episode). These relatively good-quality trajectories are put into the replay buffer $\mathcal{D}_{c}$ for updating the context encoder, which makes the sample set more complete and is conducive to correct mapping trajectories of different quality to the latent context space.

\section{Experiments}
In this section, we conducts experiments on the MuJoCo (Multi-Joint dynamics with Contact) \cite{todorov2012mujoco} and MetaWorld ML1\cite{yu2020meta} benchmark environments to verify the effectiveness of the proposed LC-SAC algorithm.
First, we briefly introduces the MuJoCo physical simulation environment and Metaworld ML1 environment, then gives the training curve of the LC-SAC and SAC algorithms, and finally studies the effect of the different hyperparameters setting of the algorithm on the performance of the algorithm.
The experimental results show that LC-SAC's performance is comparable to SAC on the continuous control benchmark task MuJoCo, and its performance is significantly better than the SAC algorithm on the MetaWorld ML1 task, which verify the importance of the context vector for solving the sequence decision-making problem in the environments with non-stationary dynamics.

Since deep reinforcement learning algorithms are highly sensitive to hyperparameters and initialization parameters, there are random and possibly unstable characteristics when training neural networks. Even for the same algorithm, due to different code implementation details and different software test environment versions, there may be big differences in performance.
To ensure the reproducibility of the experimental results, the code environment used in this paper is described as follows: OpenAI open source spinningup\footnote{https://spinningup.openai.com/en/latest/index.html} provided SAC, PPO, DDPG, TD3, etc. are used as benchmark codes, Python version is 3.6.10, PyTorch version is 1.5.0, MuJoCo and MetaWorld ML1 are selected as the test environment, OpenAI Gym is used as the application program interface, and the Gym version is 0.10.9 .
The state and action dimensions of MuJoCo and Metaworld ML1 physical simulation environment are shown in Table 1. The schematic diagram of its physical simulation environment can be found in Figure 3 and Figure 4.
The code of the experiment in this paper has been open source \footnote{https://github.com/puyuan1996/LC-SAC}

\begin{table}[!htbp]
\centering
\caption{MuJoCo physical simulation environment parameters}
\begin{tabular}{llllll}
\hline
Environment & State Dimension & Action Dimension \\
\hline
Ant-v2 & 111 & 8\\
HalfCheetah-v2 & 17 & 6\\
Hopper-v2 & 11 & 3\\
Humanoid-v2 & 376 & 17 \\
MetaWorld ML1 & 12 & 4 \\
\hline
\end{tabular}
\label{tab:booktabs}
\end{table}

\subsection{Experiment Settings}
{\bf MuJoCo}\cite{todorov2012mujoco} is a robot physical simulation environment developed by Emo Todorov for Roboti LLC. Initially, it was only used in the motion control laboratory of the University of Washington, but now it has been widely used by researchers and developers in the field of robotics and reinforcement learning.
MuJoCo includes Hopper-v2, HalfCheetah-v2, Ant-v2, Humanoid-v2 and other different tasks, the purpose of the reinforcement learning algorithm is to output the continuous control parameters of each joint of the robot, so that the robot can move as smoothly and quickly as possible. Its API is based on OpenAI's Gym \cite{brockman2016openai}, through the env.reset(), env.step() and other functions to control the simulation process. 
The specific description of the return value are shown in Table 2.
The MuJoCo environment has many different tasks with different difficulties, and the default setting of the maximum length of one episode is also different. For the 4 environments used in this paper, the maximum length of one episode is 1000. 
In addition, we used the original reward value returned by the environment and does not have special processing to the reward.

\begin{table}[!htbp]
\centering
\caption{Gym API  parameter description}
 \scalebox{0.6}{
\begin{tabular}{llllll}
\hline
Return value name & meaning \\
\hline
observation& returns the current observation of the agent \\
reward & the reward obtained by the agent, \\
& its scale range is related to the specific environment\\
 done & when it is False, it means the game is not over, when it is True, it means the game is over\\
info & Including some more detailed descriptions about the state of the agent and the environment \\
\hline
\end{tabular}
}
\label{tab:booktabs}
\end{table}

\begin{figure}[h]
\centering
\includegraphics[width=8cm,height=2.5cm]{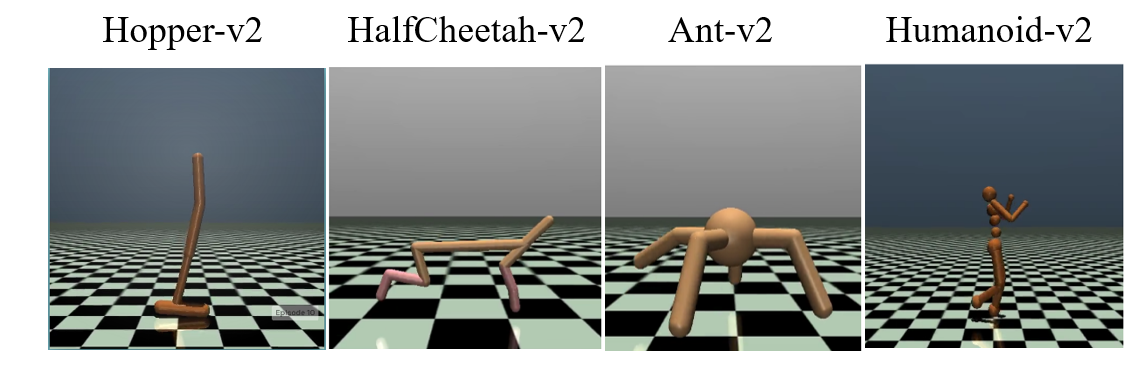}
\caption{MuJoCo physical simulation environment diagram}
\end{figure}

{\bf MetaWorld ML1} is the environment that encapsulate another layer based on MuJoCo. by Levine et al.. The agent tries to output the torque size and direction of each joint of the robotic arm in the environment, so that the robotic arm reaches, pushes or picks-places the object to a variable target Location. Including the following 3 environments:
reach-v2: Move the robot arm to the target position (shown by the red dot in the figure);
push-v2: Push the hockey puck to the target position with a robotic arm (shown by the green dot in the figure);
pick-place-v2: Use the robotic arm to hold the hockey puck and move it to the target position (shown by the blue dot in the figure).
The observation dimensions of these environments are all 12, and the action dimensions are all 4. The initial positions of these environments change in different episodes, but the target position remains unchanged.
In the following, if we add -10 after the name of the environment, such as push-v2-10, it means that 
besides the initial position in different episodes changes, the target position will also change. The number 10 after the symbol-  means there are 10 different targets position. There are also a v1 version of the MetaWorld ML1 environment. The basic principles are similar, except that the rewards scale are larger than the v2 version.

\begin{figure}[h]
\centering
\includegraphics[width=5cm,height=4cm]{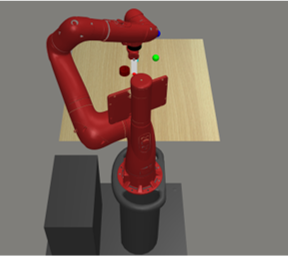}
\caption{MetaWorld ML1 physical simulation environment diagram}
\end{figure}

SAC, as the base algorithm of the LC-SAC, adopts the implementation of spinningup repository. Value function and policy function are both modeled as a feedforward fully connected network with 2 hidden layers. Each hidden layer has 256 nodes and take ReLU as the activation function.
Like the SAC algorithm, the policy in the LC-SAC algorithm is also representd by the Gaussian distribution and its covariance matrix is a diagonal matrix.
Specifically, the input of policy network is $(s_t,c_t)$, output is the mean and standard deviation of the Gaussian distribution, and during training we obtain the executed continuous action vector by sampling from the Gaussian distribution.
In the test phase, same as the SAC, LC-SAC directly executes the obtained mean vector of the Gaussian distribution. 
The value network also uses two hidden layers of multi-layer perceptrons, each hidden layer is composed of 256 nodes, the input is the splicing of the state and the context vector, and the output is a scalar, which means the estimated value of the current state-action pair.
The context encoder adopts the LSTM recurrent neural network structure. At each time t, input a triplet consisting of $(s_t,a_t,r_t)$, first pass a 128-dimensional linear layer with a hidden layer, and then pass the ReLU activation function, then pass through the 128-dimensional LSTM with a hidden layer, and finally pass through two hidden layers with a 128-dimensional linear layer, and output the context variable.
The experimental results of the PPO, DDPG, and TD3 algorithms are also obtained from running of the code implementation of spinningup, and their network structure and hyperparameters adopt the default values. The other hyperparameter settings of the LC-SAC algorithm are shown in Table 3.

\begin{table}[!htbp]
\centering
\caption{hyper-parameter setting of LC-SAC algorithm}
    \scalebox{0.6}{
    \begin{tabular}{ll}
    \hline
    Parameter name & value \\
    \hline
    Learning rate & 3e-4 \\
    Target network smoothing coefficient ($\tau$) & 0.005\\
    Entropy coefficient ($\alpha$) & 0.2 \\
    Discount factor & 0.99 \\
    latent context variable dimension & 5 \\
    latent context variable encoder network structure & Linear-Linear (latent dim=256) \\
    Value function and policy function network structure & Linear-LSTM-Linear-Linear(latent dim=128) \\
    Optimizer & Adam \cite{kingma2014adam} \\
    Activation function & ReLU \\
    replay buffer size $rlbs$ & 1e6 \\
    Update start time step & 2000\\
    The mini-batch size used to update the network $l$ & 128 \\
    Length of continuous sequence segment of trajectory   &20\\
    KL difference (Kullback-Leibler divergence) weight lambda & 0.2 \\
    \hline
    \end{tabular}
    }
\label{tab:booktabs}
\end{table}


\subsection{Experimental Results on MuJoCo}
\begin{figure*}[htbp]
\centering
\includegraphics[width=14cm,height=6cm]{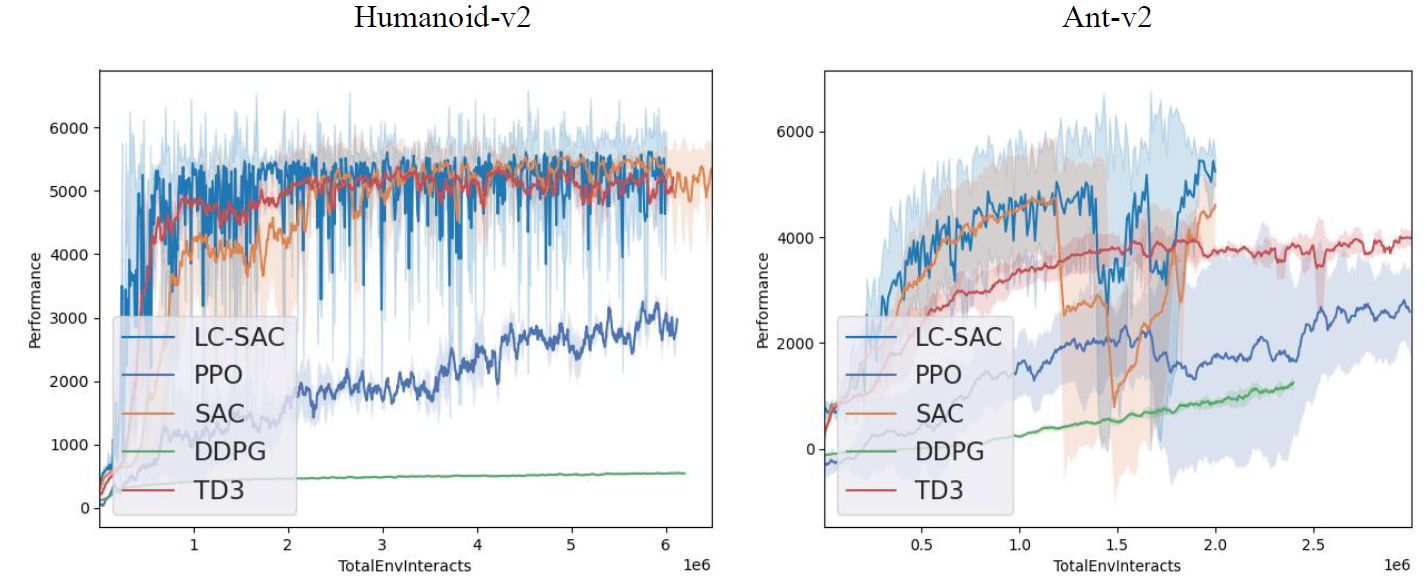}
\includegraphics[width=14cm,height=6cm]{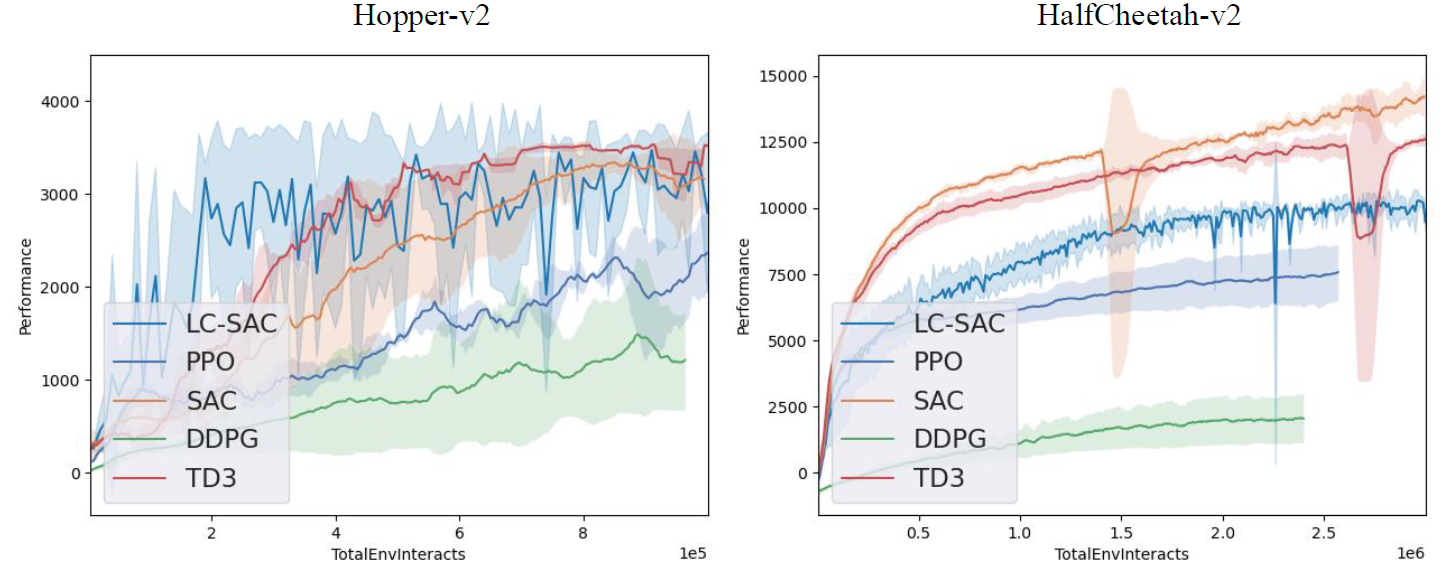}
\caption{Training curves of different algorithms on different tasks of MuJoCo}
\label{wolf2}
\end{figure*}
On continuous control tasks in different environments of MuJoCo, run LC-SAC, SAC, and other mainstream reinforcement learning algorithms like PPO, DDPG, TD3, and draw their test scores graph over time.
Figure 5 shows how the average test score of each method in 5 runs varies with the number of interactions with the environment on different tasks. The scale of the horizontal axis is 1e6 or 1e5 time steps. In each run of the algorithm, the training of the neural network is suspended every 10,000 steps, 
the agent execute the deterministic mean action and interact with the environment. The cumulative returns are averaged under 10 different random seeds.
The shaded area represents one-half of the standard deviation of the scores obtained from 5 runs. The blue curve represents the LC-SAC algorithm, the orange curve represents the SAC algorithm, the abscissa represents the number of interactions with the environment, and the ordinate represents the scores of different algorithms under the number of environmental interactions corresponding to the abscissa.
It is worth noting that, since these tasks in the MuJoCo environment may essentially satisfy the assumptions of the Markov decision process, the experiment in this subsection mainly illustrates the convergence and effectiveness of the LC-SAC algorithm, which is mainly applicable to the environments with non-stationary dynamics described in the section.

According to the experimental results, it can be seen that the LC-SAC has achieved similar results to SAC on Hopper-v2 and learns faster in Humanoid-v2 and Ant-v2, but on relatively simple tasks such as HalfCheetah-v2, the performance of LC-SAC is slightly worse.
On tasks like HalfCheetah-v2, the optimal action of the agent at each time step may only depend on the current state observations. The introduction of latent context variables does not help much for policy optimization, but increases the computational overhead and decrease the asymptotic performance
On tasks such as Humanoid-v2, the decision at certain time step needs to take into account the recent information. The learned latent context vector not only contains the dynamic information of the environment, but also the behavior information of the agent in the recent period, so, in the result, the agent of the LC-SAC algorithm learns slightly faster than the SAC algorithm.
\subsection{Experimental Results on MetawWorld ML1}

\begin{figure*}[htbp]
\centering
\subfigure{
\includegraphics[width=0.4\textwidth]{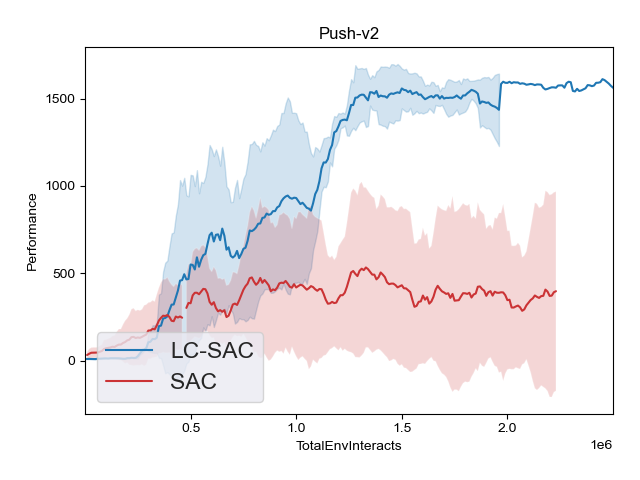}
}
\subfigure{
\includegraphics[width=0.4\textwidth]{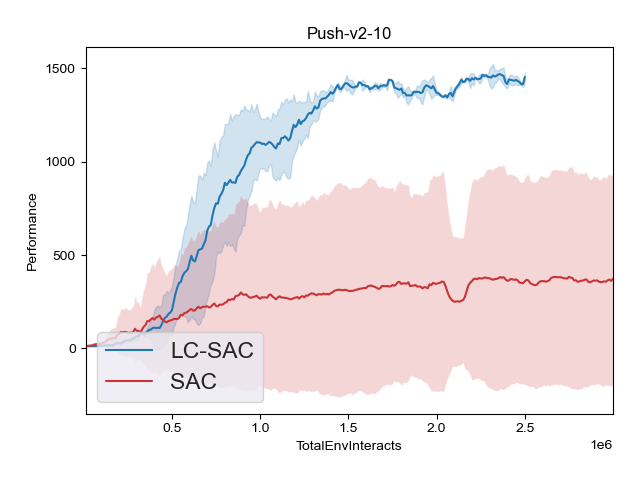}
}
\subfigure{
\includegraphics[width=0.4\textwidth]{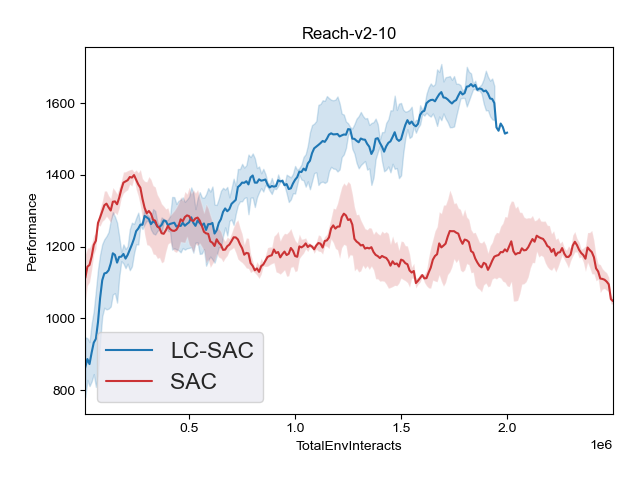}
}
\subfigure{
\includegraphics[width=0.4\textwidth]{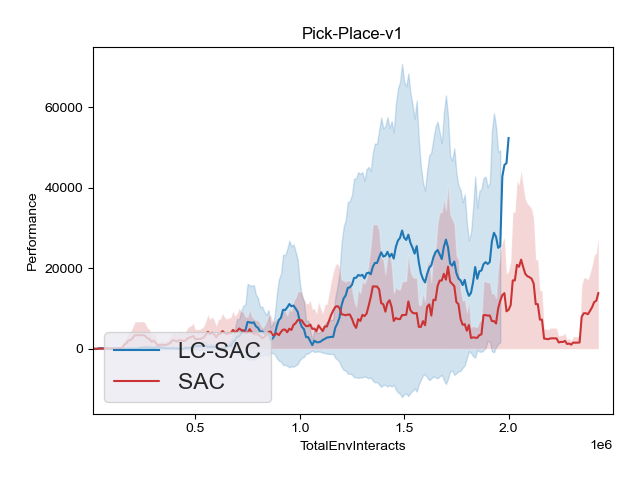}
}
\caption{On different tasks of MetaWorld ML1, LC-SAC and SAC algorithm training graphs}
\label{wolf2}
\end{figure*}

This subsection mainly conducts experiments on the MetaWorld ML1 environment where the dynamics changes significantly than MuJoCo in the last subsection. In addition, comparative experiments are carried out on different hyperparameter settings in the LC-SAC algorithm. After analyzing the experimental results, we give some reasonable suggestions for hyperparameter settings.
The meaning of the graph and the test method of algorithm performance are basically the same as in the previous subsection, so we won't go into details here.

First of all, under the default hyperparameter settings, that is, the context variable is a deterministic value, the dimension of context variable is 5, and the length of sampled trajectory segment for updating context encoder is 20, and the loss function for updating context encoder only contains the contrastive prediction loss function. Run LC-SAC and SAC algorithms on different tasks (Push-v2-1, Push-v2-10, Reach-v2-10) in MetaWorld ML1 environment, and graph the test scores over time 
, as shown in Figure 6. The blue curve represents the LC-SAC algorithm.
From the experimental results, it can be seen that the LC-SAC algorithm 
 significantly outperforms the original SAC algorithm. The original SAC algorithm assumes that the underlying dynamics of environment is a fixed Markov decision process, but the actual dynamics of different tasks in the MetaWorld ML1 vary from episode to episode.
 As a result, the optimality of the original SAC algorithm cannot be guaranteed
   
In the rest of this subsection, we studied the effect of different hyperparameter settings on the performance of the LC-SAC algorithm through experiments. HyperparameterS including the update frequency of latent context variables $N_c$, the size of the reinforcement learning replay buffer $rlbs$, whether the latent context variable is a probability value, and whether the loss function contains Q loss function 

After analyzing the experimental results, the following hyperparameter setting suggestions are given:

{\bf Setting the context variable to a deterministic value is conducive to the rapid improvement of the policy}. The performance of the LC-SAC algorithm when the context variable is deterministic and a probabilistic was tested in the Push-v2-1 environment, 
, the experimental results are shown in Figure 7.
The $sar$ in the legend indicates that the observation data $e$ at the next time step is the stitching vector of $(s_{t+1},a_{t+1},r_{t+1})$, the $s$ in the legend means that the observation data at the next time step is directly taken as the state $s_t$ at the next time step, the $det$ means the context variable takes a deterministic value, and $prob $ means the context variable takes the deterministic value.
The results show that the selection of the context vector as a deterministic form will obviously benefit the rapid and stable improvement of the policy, and if context vector is probability distribution form, the learning process will often slow down a lot, 
and for the algorithm that uses probabilistic context vectors,
more training samples are required to achieve the same performance as the algorithm that uses deterministic context vectors.
In the context encoder loss function, e is setted as the next state $s_t$ or the concatenation vector of $(s_{t+1},a_{t+1},r_{t+1})$, corresponding performance of the algorithm is similar, but theoretically, only if we can  predict the transition $(s_{t+1},a_{t+1},r_{t+1})$ at next time step from the context vector $c_t$, the $c_t $ contains dynamics information of the environment,
Therefore, in all later experiments of this paper, unless specified, the context vector adopted deterministic value and the observation data $e$ at the next time step in the context loss function is
concatenation vector of $(s_{t+1},a_{t+1},r_{t+1})$.
\begin{figure}[htbp]
\centering
\subfigure{
\includegraphics[width=0.4\textwidth]{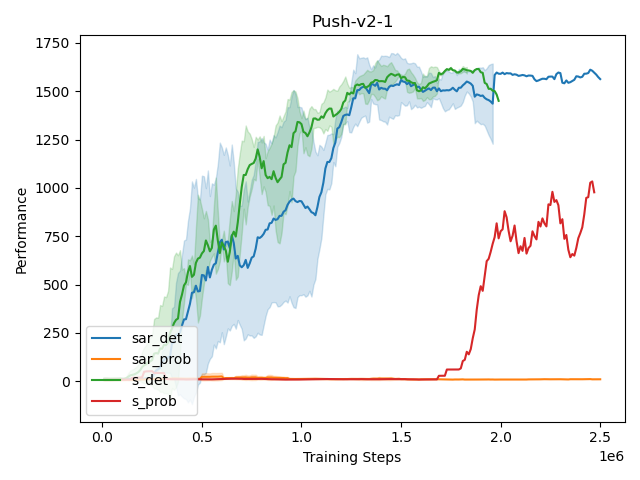}
}
\caption{The context variable is set to the deterministic value or the probability value, and
the observation data $e$ at the next time step in the context loss function is
concatenation vector of $(s_{t+1},a_{t+1},r_{t+1})$ or only the state at the next time step $s_t$}
\label{wolf2}
\end{figure}

{\bf The size of the replay buffer should be appropriately set to be larger}. In the Reach-v2-10 environment, the performance of the LC-SAC algorithm was tested when the length of the replay buffer used for updating the value function network and the policy network is 1e6 and 1e5 respectively, as shown in Figure 8.
The experimental results showed that when the size of the replay buffer used to update the value network and the policy network is reduced, the performance of the LC-SAC algorithm will be significantly decreased.
\begin{figure}[htbp]
\centering
\subfigure{
\includegraphics[width=0.4\textwidth]{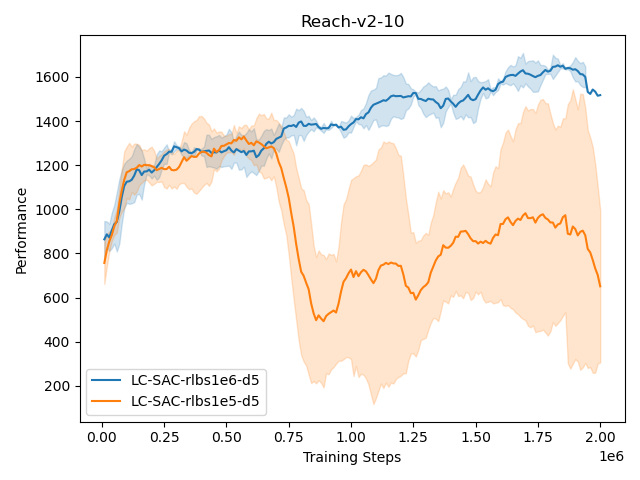}
}
\caption{Training curve graph when the length of the replay buffer used for updating the value network and the policy network is 1e6 and 1e5 respectively}
\label{wolf2}
\end{figure}

\begin{figure}[!htbp]
\centering
\subfigure{
\includegraphics[width=0.4\textwidth]{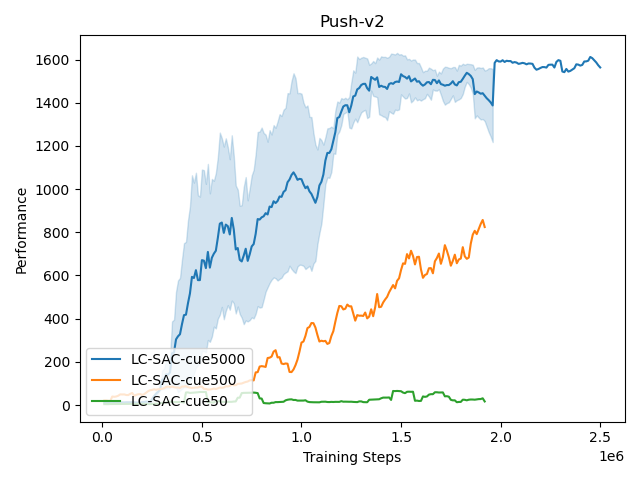}
}
\caption{Training curve graph when the update frequency of the context encoder $N_c$ is set to different values}
\label{wolf2}
\end{figure}

{\bf Compared with the parameter update frequency of policy network and value network $N_{rl}$, the parameter update frequency $N_c$ of the context encoder should be set slightly larger}. The performance of the LC-SAC algorithm when the update frequency of the latent context vector $N_c$ is 5000, 500, and 50 is tested in the Push-v2 environment, as shown in Figure 9.
The number after $cue$ in the legend indicates the parameter update frequency of the context encoder $N_c$. For example, $cue5000$ means that the context encoder is optimized every 5000 steps of interaction with the environment. Note that the parameter update frequency of the policy network and the value network $N_{rl}$ is consistent with the original SAC algorithm, and the default is 50.
The experimental results indicates that making the update frequency of the context vector $N_c$ greater than the update frequency of the policy network and the value network $N_{rl}$=50 will make the algorithm performance more robust.


{\bf Including the Q loss function in the loss function of updating context encoder is not necessarily conducive to the rapid improvement of the policy}. The performance of the LC-SAC algorithm when the coefficients corresponding to $L_critic$ in the loss function of updating context encoder are 0, 0.5, and 1, respectively, are tested in the Push-v1-1 environment, as shown in Figure 10. The results show that when the coefficients corresponding to the Q loss function have different values, the final test scores are not much different. It can be seen that including the Q loss function in the loss function of  updating context encoder is not necessarily conducive to the rapid improvement of the policy.
\begin{figure}[!htbp]
\centering
\subfigure{
\includegraphics[width=0.4\textwidth]{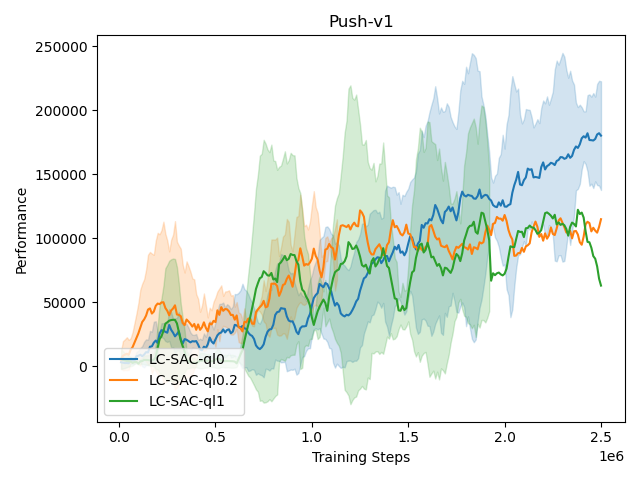}
}
\caption{The training curve where the coefficients corresponding to the $L_{critic}$ item in the loss function of updating context encoder are 0, 0.5 and 1 respectively}
\label{wolf2}
\end{figure}

\begin{figure}[!htbp]
\centering
\subfigure{
\includegraphics[width=0.4\textwidth]{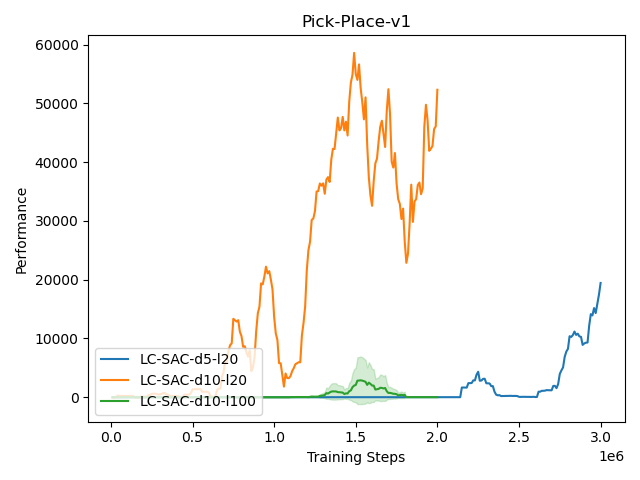}
}
\caption{Training curve for different context vector dimensions and different length of the trajectory segment used updating context encoder}
\label{wolf2}
\end{figure}

{\bf On relatively complex tasks, the dimension of the context variable needs to be appropriately set to be larger}. In the Pick-Place-v1-1 environment, we tested the performance of the LC-SAC algorithm when the context vector dimensions are 5 and 10, and the length of the trajectory segment used to update context encoder are 20 and 100, as shown in Figure 11. 
The results show that for the task Pick-Place-v1-1, the dimension of context vector set as 10 is better than 5.
But the length of the trajectory segment used to update context encoder set as 20 is better than 100
indicating that the length should not be too long, because when calculating the contrastive prediction loss function, we only use the context variable at the last time step of the sampled sequence and our purpose is to make the learned context vector capture the dynamics of environment. If the sequence length is too long, it will increase the calculation overhead, resulting in poor algorithm sample efficiency, and is not necessary for context vector capturing the dynamics of environment.

\section{Conclusion and Future Works}
In this paper, we proposes a context-based soft actor critic method (LC-SAC) for environments with non-stationary dynamics. LC-SAC introduces a latent context encoder utilizing the recurrent neural network, and minimizes the contrastive prediction loss function to
maximize the mutual information between the context variable with the transition in the future, thus implicitly prompts the latent context variable to contain information about the environment dynamics and the recent behavior of the agent.
At each time step, by directly concatenating the learned context vector with the original state and use it as the new input of the value function network and the policy network,
then alternately optimize the value network and policy network parameters in the soft policy iteration paradigm, and finally approach the optimal policy.

In the experimental part, firstly, experiments were carried out on different tasks in the continuous control benchmark MuJoCo where the dynamic changes were not so obvious. The results showed that the performance of LC-SAC was comparable to that of SAC, which verified the convergence of the LC-SAC algorithm.  On the MetaWorld ML1 task, the performance of the LC-SAC algorithm is significantly better than that of the SAC algorithm, which verifies the importance of the latent context vector for solving the sequence decision problem in the environment with non-stationary dynamics. 
In addition, we conduct experements on the effects of various hyperparameters in the LC-SAC algorithm on the performance of the algorithm and at last we give some reasonable hyperparameter setting suggestions.





\vspace{12pt}

\end{document}